%% file: main.tex
\documentclass[10pt,twocolumn,letterpaper]{article}

\usepackage[pagenumbers]{cvpr}

\usepackage{graphicx}
\usepackage{booktabs}
\usepackage{multirow}
\usepackage{microtype}
\usepackage{xspace}
\usepackage{xcolor}
\usepackage[sort&compress]{natbib}

\input{preamble}

\definecolor{cvprblue}{rgb}{0.21,0.49,0.74}
\usepackage[pagebackref,breaklinks,colorlinks,allcolors=cvprblue]{hyperref}

\def\paperID{0000}
\def\confName{arXiv}
\def\confYear{2026}

\title{TriWorldBench: A Tri-View Consistency
Perspective on Embodied World Models}

\author{
Xuanyi Liu\textsuperscript{1} \quad
Haofeng Wang\textsuperscript{1} \quad
Ruiqi Li\textsuperscript{1} \quad
Danni Yu\textsuperscript{2} \quad
Rui Wan\textsuperscript{2} \quad
Ruixu Zhang\textsuperscript{2} \quad
Siyu Tao\textsuperscript{3} \quad \\
Xue Yang\textsuperscript{4} \quad
Shaofeng Zhang\textsuperscript{5} \quad
Zicheng Zhang\textsuperscript{6} \quad
Jiaqi Zhang\textsuperscript{1} \quad
Siwei Ma\textsuperscript{1} \\[6pt]
{\normalsize
\textsuperscript{1}Peking University \quad
\textsuperscript{2}Tsinghua University \quad
\textsuperscript{3}Beihang University \quad
\textsuperscript{4}Shanghai Jiao Tong University} \\
{\normalsize
\textsuperscript{5}University of Science and Technology of China \quad
\textsuperscript{6}Shanghai AI Laboratory}
}

\begin{document}
\maketitle

\input{sec3/0_abstract}
\input{sec3/1_intro}
\input{sec3/2_related}
\input{sec3/3_benchmark}
\input{sec3/4_protocol}
\input{sec3/5_conclusion}

\clearpage
{
    \footnotesize
    \bibliographystyle{ieeenat_fullname}
    \bibliography{main}
}

\end{document}

%% file: preamble.tex
\newcommand{\method}{\textsc{TriWorldBench}\xspace}
\newcommand{\score}{\mbox{TWB-Score}\xspace}

\usepackage{needspace}

\usepackage{etoolbox}
\makeatletter
\patchcmd{\@maketitle}{\vskip .375in}{\vskip .25in}{}{%
  \PackageWarningNoLine{preamble}{could not tighten title lead-in}}
\patchcmd{\@maketitle}{\vspace*{24pt}}{\vspace*{14pt}}{}{%
  \PackageWarningNoLine{preamble}{could not tighten space below title}}
\patchcmd{\@maketitle}{\vspace*{12pt}}{\vspace*{6pt}}{}{%
  \PackageWarningNoLine{preamble}{could not tighten space below authors}}
\makeatother

\newcommand{\projectcode}{\href{https://github.com/TriWorldBench/TriWorldBench}%
  {https://github.com/TriWorldBench/TriWorldBench}}

%% file: sec3/0_abstract.tex
\begin{abstract}
Embodied world models predict the outcomes of robot actions to support learning
and planning. For robots equipped with head and wrist cameras, this requires
complementary views: the head view captures the overall task, while wrist views
reveal local gripper--object interactions. However, evaluating these
views independently cannot determine whether they describe the same action and
object state. We introduce \method, a benchmark for evaluating embodied world
models through synchronized head, left-wrist, and right-wrist videos. It contains
500 episodes across 50 bimanual manipulation tasks and uses 19 metrics to assess
tri-view consistency, task alignment, physical and 3D coherence, motion quality,
temporal consistency, and visual quality. By combining cross-view checks with
measurements tailored to each camera, the benchmark evaluates whether plausible
individual videos also form a consistent prediction of the intended task.
We summarize overall performance with \score and retain per-view results to
identify where predictions fail. This extends world-model evaluation beyond
single-view visual quality.
Code, data, and metric definitions are available at \projectcode.
\end{abstract}

%% file: sec3/1_intro.tex
\section{Introduction}
\label{sec:intro}

A world model predicts how an environment evolves and can support robot learning
and planning~\citep{ha2018worldmodels,yang2024unisim,assran2025vjepa2}.
Action-conditioned video models make these predictions available as synthetic
training trajectories, policy rollouts, or visual subgoals~\citep{guo2026ctrlworld}.
These uses require more than convincing images: a predicted grasp that looks
plausible but violates object contact can provide misleading supervision or
guide a planner toward an infeasible action. Evaluation must therefore connect
visual quality to the manipulation event a video claims to predict.

\begin{figure}[t]
    \centering
    \includegraphics[width=\linewidth]{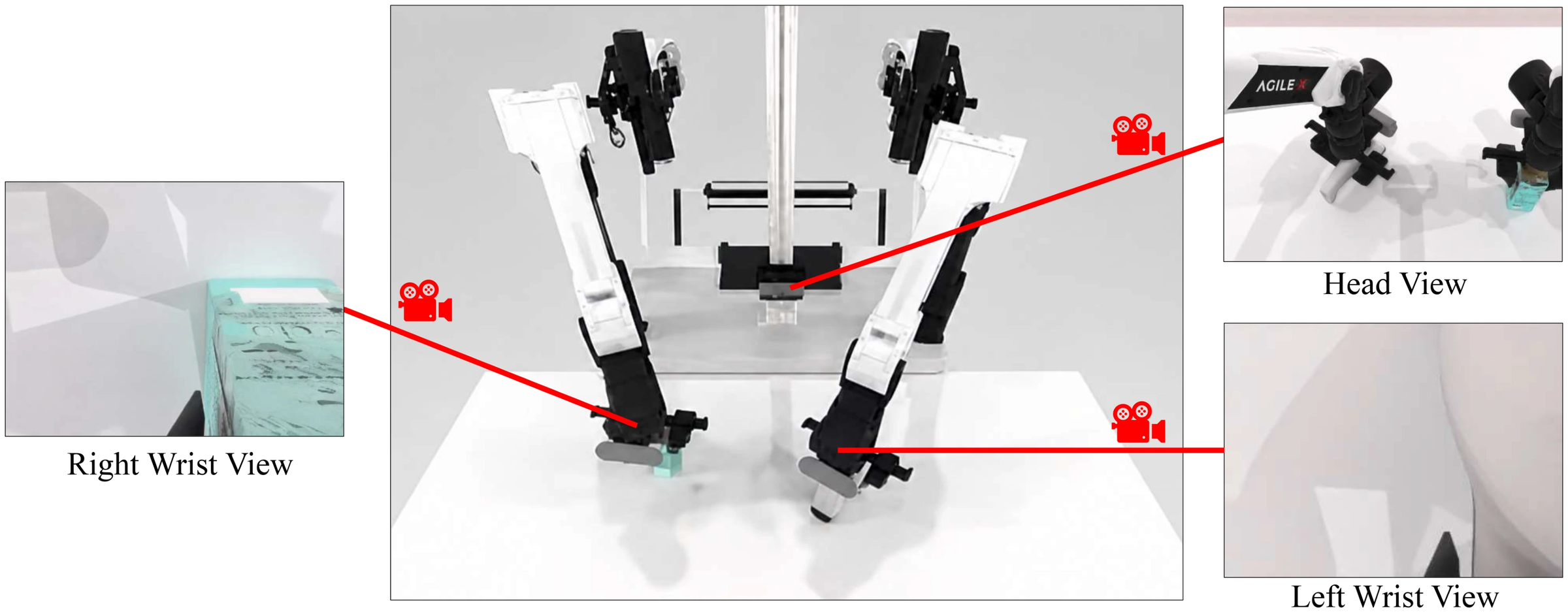}
    \caption{Complementary views of one manipulation event. The head camera
    provides workspace context; the wrist cameras reveal local geometry and
    gripper--object contact. \method evaluates both the quality of each
    predicted view and agreement across the three views.}
    \label{fig:cameras}
\end{figure}

The camera setup changes what this evaluation must establish.
In the head-and-wrist configuration of Figure~\ref{fig:cameras}, the head camera
provides workspace context and the wrist cameras reveal local geometry and
gripper--object contact~\citep{hsu2022hands,khazatsky2024droid}. Their evidence
is complementary, not interchangeable: a slipped grasp may be visible only at a
wrist, while a close-up may hide which object or arm the task requires.
Recent world models predict multiple camera streams
jointly~\citep{guo2026ctrlworld,liu2026geometry4d}. Yet even three individually
plausible videos may disagree about whether an object is held, released, or
transferred. Single-view quality cannot establish this shared
state~\citep{shi2024mvdream,liu2024syncdreamer}. Conversely, different
appearances do not necessarily imply inconsistency: wide camera baselines,
wrist motion, and occlusion naturally change what is visible. The relevant
question is whether all views are compatible with the same event, not whether
they look alike.

\method makes this distinction explicit by treating a synchronized triplet as
one prediction. Building on embodied world-model
evaluation~\citep{li2025worldmodelbench,yue2025ewmbench,%
qin2024worldsimbench,shang2026worldarena}, it asks three complementary questions:
is each view plausible, do the views agree, and does the prediction match the
task? Cross-view consistency is a dedicated dimension, not a substitute for
task or visual quality. Metrics use the cameras best suited to their evidence,
and reference action phases distinguish expected stillness from missing motion.
The contribution is thus to assess the relationship between views while
preserving the different information each camera provides.
This helps distinguish predictions that merely look coherent from those whose
views jointly support the intended robot behavior.
Table~\ref{tab:metrics} summarizes the 19 metrics and the camera views used
to compute each score.

\begin{table*}[t]
    \centering
    \scriptsize
    \renewcommand{\arraystretch}{0.9}
    \setlength{\tabcolsep}{5pt}
    \caption{The 19 metrics reported by \method, grouped into six dimensions.
    H, L, and R denote the head, left-wrist, and right-wrist cameras;
    ``H\,+\,wrist'' denotes a comparison between the head view and the wrist view
    of the acting arm; ``joint'' denotes a measurement over the triplet as a
    whole. H/L/R metrics are computed separately in each view before
    aggregation. Pixel and structural comparisons use the corresponding
    ground-truth camera, not a different viewpoint.}
    \label{tab:metrics}
    \begin{tabular}{@{}p{0.135\textwidth}p{0.145\textwidth}p{0.58\textwidth}l@{}}
        \toprule
        Dimension & Metric & What it measures & Cameras \\
        \midrule
        Tri-view
        consistency
          & Normalized PSNR   & Pixel-level agreement with the corresponding ground-truth camera & H/L/R \\
          & SSIM              & Structural similarity to the corresponding ground-truth camera & H/L/R \\
          & VLM Consistency I   & Compatibility-first head-to-wrist check of robot state, phase, and object state; occluded evidence tolerated & H\,+\,wrist \\
          & VLM Consistency II  & Verification-first check: criteria not verifiable from both views receive a deduction & H\,+\,wrist \\
          & VLM Consistency III & Exemplar-calibrated against labeled cases of blur, wrong held object, and wrong acting arm & H\,+\,wrist \\
          & VQA Consistency   & Accuracy on a question bank frozen in advance from the reference triplet and phases & joint \\
        \midrule
        Task
        alignment
          & Instruction Following & Whether the rollout uses the required arm and object and reaches the goal state & H \\
          & Semantic Alignment    & Similarity between generated and reference head-video captions in text embedding space & H \\
          & JEPA Similarity       & Feature similarity to the reference triplet under a frozen video encoder, camera order preserved & joint \\
        \midrule
        Physical and\newline
        3D coherence
          & Interaction Quality & Plausibility of contact, grasp stability, object response, and freedom from penetration & H/L/R \\
          & Perspective         & Depth-consistent scale, occlusion ordering, shape stability, and camera geometry & H/L/R \\
        \midrule
        Motion
        quality
          & State Alignment      & Agreement with the moving or static pattern expected for each view in the reference phase & H/L/R \\
          & Flow Score           & Optical-flow magnitude~\citep{teed2020raft}, indicating visible motion in each view & H/L/R \\
          & Trajectory Accuracy  & Agreement of predicted and reference end-effector paths under time warping & H \\
        \midrule
        Temporal
        consistency
          & Subject Consistency     & Stability of robot and object identity and appearance across frames & H/L/R \\
          & Background Consistency  & Stability of background features across frames & H/L/R \\
          & Photometric Smoothness  & Motion-aligned appearance stability; penalizes flicker and texture drift & H/L/R \\
        \midrule
        Visual
        quality
          & Image Quality     & Frame clarity: blur, noise, exposure, compression~\citep{ke2021musiq} & H/L/R \\
          & Aesthetic Quality & Frame-level visual appeal, reported separately from task correctness & H/L/R \\
        \bottomrule
    \end{tabular}
\end{table*}

%% file: sec3/2_related.tex
\section{Related Work}
\label{sec:related}

\paragraph{Embodied world models.}
World models support robot learning and
planning~\citep{yang2024unisim,assran2025vjepa2}. EnerVerse-AC, Genie Envisioner,
DreamDojo, and Motus predict action-conditioned robot
video~\citep{jiang2025enerverseac,liao2025genieenvisioner,%
gao2026dreamdojo,bi2025motus}. PAVXploreRL further uses physical, action, and
visual criteria as post-training rewards~\citep{wang2026pavxplorerl}.
As predictions become training data and decision inputs, evaluation must assess
task behavior and agreement across cameras.

\paragraph{Multi-view prediction and consistency.}
Multi-view models explicitly represent camera relationships. Ctrl-World predicts
external and wrist views
jointly~\citep{guo2026ctrlworld}, geometry-aware video generation aligns views
through point maps~\citep{liu2026geometry4d}, and ReViWo learns representations
robust to camera disturbance~\citep{pang2025reviwo}. In image and video generation,
shared priors, 3D-aware attention, and geometric distillation promote geometric
consistency~\citep{shi2024mvdream,liu2024syncdreamer,liu2026camgeo}. \method tests whether such
predictions describe compatible robot and object states despite motion and occlusion.

\paragraph{Video and world-model benchmarks.}
Benchmarks distinguish visual fidelity from useful prediction. VBench evaluates
perceptual and temporal
dimensions~\citep{huang2024vbench}; WorldScore evaluates controllability, quality,
and dynamics~\citep{duan2025worldscore}. WorldModelBench tests instruction
following and physical laws~\citep{li2025worldmodelbench}, while EWMBench
separates scene consistency, motion correctness, and semantic
alignment~\citep{yue2025ewmbench}. WorldSimBench combines human-aligned perceptual
judgments with action-level evaluation~\citep{qin2024worldsimbench}. WorldArena
further connects perceptual assessment to functional utility, evaluating world
models as data engines, policy evaluators, and action
planners~\citep{shang2026worldarena}. This motivates separating visual appeal
from task alignment. RBench evaluates robot-oriented task correctness and physical
plausibility~\citep{deng2026rbench}, while RoboWM-Bench tests whether generated
behaviors translate into executable manipulation~\citep{jiang2026robowmbench}.
\method adapts these dimensions to complementary camera roles and explicitly
evaluates agreement between synchronized views.

%% file: sec3/3_benchmark.tex
\section{Benchmark Overview}
\label{sec:benchmark}

\paragraph{Task and data.}
Each episode provides three synchronized initial frames, a natural-language
instruction, and an action sequence. A model predicts a head-view video and two
wrist-view videos covering the same time span; corresponding timestamps denote
the same instant of the task. The test set contains 500 episodes spanning the
50 bimanual manipulation tasks of RoboTwin~2.0~\citep{chen2025robotwin}, including
pick-and-place, stacking, pouring, handover, switch actuation, and tool use.
Tasks are recorded under clean and domain-randomized background conditions,
introducing appearance variation alongside task diversity. A separate
100-episode validation split with reference videos and state annotations supports
local development.\footnote{\url{https://huggingface.co/datasets/TriWorldBench/Dataset}}

\paragraph{Reference-derived action states.}
Reference states provide a shared temporal context for evaluation. Fixed rules
over gripper opening and end-effector displacement segment each reference
trajectory into coarse phases, including idle, approach, grasp, hold,
manipulation, release, and completion. The annotations identify the acting arm
and support the expected moving or static pattern for each view. They also
select phase-level keyframes and the active head-to-wrist pair
(Figure~\ref{fig:state}). These are heuristic action labels, not verified
task-success labels. Derived only from the reference trajectory, the boundaries
remain fixed across candidate models, so a model's own failure cannot redefine
the phase against which it is judged.

%% file: sec3/4_protocol.tex
\section{Evaluation Protocol}
\label{sec:method}

The 19 metrics in Table~\ref{tab:metrics} separate within-view quality from
cross-view agreement and task alignment. The protocol combines reference-based
comparisons with judgments of the generated videos themselves. Four design
choices adapt these measurements to synchronized robot cameras.

\paragraph{Tri-view world consistency.}
Consistency is assessed both indirectly through the ground truth and directly
across views. Normalized PSNR and SSIM compare each generated stream with its
corresponding ground-truth stream. Because the reference triplet is synchronized
and internally consistent, proximity to it provides an indirect consistency
signal. It is not proof of agreement, however, and a valid alternative execution
may differ from the reference. Complementary vision-language model (VLM)
judgments compare the head and active wrist views. Three rubrics distinguish
compatibility despite occlusion, explicit verification in both views, and
calibration against labeled examples. Reporting them separately makes their
different treatment of uncertain evidence explicit. Visual question answering
(VQA) further tests the triplet using questions fixed from the reference and its phases.
Together, these checks ask whether the views describe compatible robot and
object states without requiring identical appearance across cameras.

\paragraph{Camera-specific evidence.}
Camera selection follows what a metric needs to observe. Instruction Following
uses the head view for task context; Semantic Alignment compares VLM-generated
descriptions of the generated and reference head videos in text embedding
space. Trajectory Accuracy also uses the head view, where arm paths are more
reliably visible than in wrist close-ups. By contrast, JEPA Similarity compares
the generated and reference triplets through a frozen video encoder with camera
order preserved. Interaction Quality and Perspective are evaluated separately
in all three views, capturing local contact and geometry as well as the wider
scene. Per-view scores retain these differences instead of treating every
camera as an equally reliable observer of every property.
For example, a wrist view can reveal a slipping grasp despite plausible
head-view object motion.

\begin{figure}[t]
    \centering
    \includegraphics[width=0.82\linewidth]{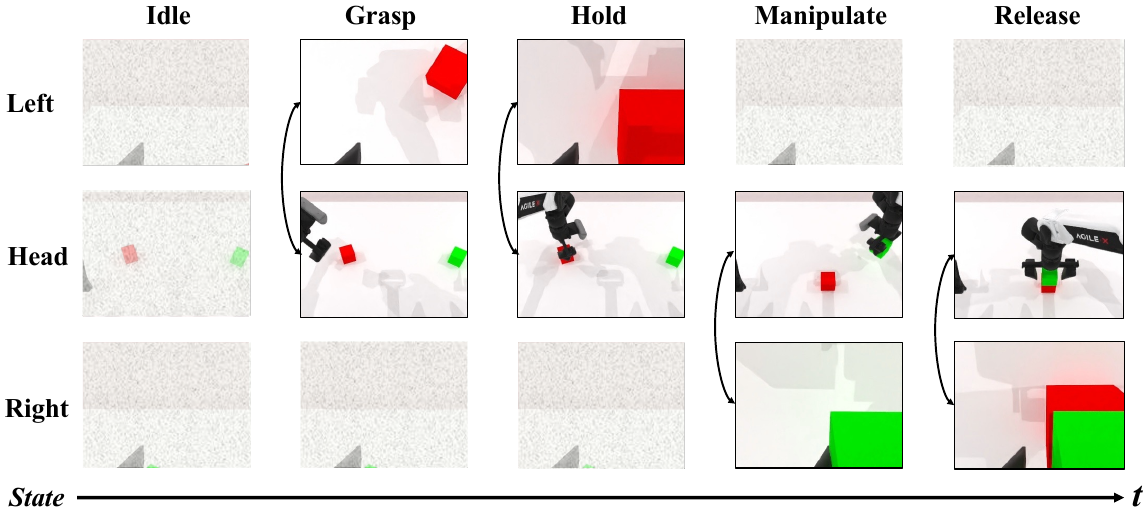}
    \caption{Reference-derived action phases select synchronized keyframes and
    head-to-wrist pairs (arcs). They also provide context for evaluating the
    expected motion in each view.}
    \label{fig:state}
\end{figure}

\paragraph{State-conditioned motion.}
Temporal stability is meaningful only relative to expected motion.
State Alignment checks each view's moving or static pattern; Flow Score measures
optical-motion magnitude; Trajectory Accuracy compares head-view end-effector
paths with the reference under time warping. More motion is not necessarily more
correct motion. Phase-conditioned penalties also adjust subject, background,
and photometric consistency when expected motion is missing or unexpected motion
appears. This discounts frozen predictions during active phases while preserving
legitimate stillness.

\paragraph{Task-grounded visual quality.}
Visual appeal and task correctness are complementary, not interchangeable.
Image and aesthetic scores are therefore accompanied by penalty-adjusted
versions that discount quality when head-view trajectory accuracy or cross-view
consistency is low. This reduces credit for attractive but unreliable
predictions while keeping raw visual quality separate from the task and
consistency evidence used to qualify it.

\paragraph{Aggregation.}
Metrics are averaged within each dimension, and the six dimension scores are
combined into \score. Dimension-level, per-view, head-to-wrist, and joint scores
remain available: the aggregate summarizes performance, while the detailed
scores help distinguish poor appearance, task mismatch, and cross-view conflict.

%% file: sec3/5_conclusion.tex
\section{Conclusion}
\label{sec:conclusion}

\method evaluates triple-view world models through the event their videos
jointly describe. Its central distinction is between a view that looks
plausible, views that agree, and a prediction that matches the task. Camera-aware
measurement and reference-derived phases make these questions explicit across
19 metrics, with \score providing a compact summary. The resulting framework
makes cross-view consistency measurable without reducing world-model quality
to consistency alone. It supports targeted diagnosis of failures in triple-view
predictions for robot learning and planning.